\pdfoutput=1

\documentclass[11pt]{article}

\usepackage[final]{acl}

\usepackage{times}
\usepackage{latexsym}
\usepackage{color}
\usepackage{graphicx}
\usepackage{multirow}
\usepackage{makecell}
\usepackage{subfigure}
\usepackage{algorithm}
\usepackage{algpseudocode}
\usepackage{booktabs} 
\usepackage{amssymb}
\usepackage{amsmath}
\usepackage{bbm}
\usepackage{xspace}
\usepackage{pifont}
\definecolor{shadecolor}{rgb}{0.92,0.92,0.92}
\usepackage{framed}

\usepackage[T1]{fontenc}

\usepackage[utf8]{inputenc}

\usepackage{microtype}

\usepackage{inconsolata}

\usepackage{graphicx}

\title{DPF-CM: A Data Processing Framework with Privacy-Preserving Vector Databases for Chinese Medical LLMs Training and Deployment}

\author{
 \textbf{Wei Huang\thanks{These authors contributed equally to this work}},
 \textbf{Anda Cheng$^*$},
 \textbf{Zhao Zhang},
 \textbf{Yinggui Wang}\thanks{Corresponding author (wyinggui@gmail.com).},
\\
 Ant Group, China
\\
   hw378176@antgroup.com,andacheng.cad@gmail.com \\
   quanjun.zz@antgroup.com, wyinggui@gmail.com
}

\begin{document}
\maketitle

\begin{abstract}

Current open-source training pipelines for Chinese medical language models predominantly emphasize optimizing training methodologies to enhance the performance of large language models (LLMs), yet lack comprehensive exploration into training data processing. To address this gap, 
we propose \textbf{DPF-CM}, a holistic \textbf{D}ata \textbf{P}rocessing \textbf{F}ramework for \textbf{C}hinese \textbf{M}edical LLMs training and deployment.
DPF-CM comprises two core modules. 
The first module is a data processing pipeline tailored for model training. 
Beyond standard data processing operations, we
(1) introduce a \textit{chained examples context-learning strategy} to generate question-oriented instructions to mitigate the lack of instruction content, and
(2) implement an \textit{ensemble-based filtering mechanism} for preference data curation that averages multiple reward models to suppress noisy samples.
The second module focuses on privacy preservation during model deployment. 
To prevent privacy risks from the inadvertent exposure of training data, we propose a \textbf{P}rivacy \textbf{P}reserving \textbf{V}ector \textbf{D}atabase (\textbf{PPVD}) approach, 
which involves model memory search, high-risk database construction, secure database construction, and match-and-replace, four key stages to minimize privacy leakage during inference collectively.
Experimental results show that DPF-CM significantly improves model accuracy, enabling our trained Chinese medical LLM to achieve state-of-the-art performance among open-source counterparts. Moreover, the framework reduces training data privacy leakage by 27\%.

\end{abstract}

\section{Introduction}

Recent large language models (LLMs) have achieved remarkable breakthroughs and are capable of answering a wide range of questions~\cite{achiam2023gpt,wang2023privacy}. However, while some models perform well in Chinese tasks, they still struggle in specialized domains such as Chinese healthcare due to limited professional knowledge~\cite{zhao2023survey, huang2024fast}. Moreover, models like GPT-4, although providing detailed responses, often lack the interactive and diagnostic capabilities typical of doctors~\cite{zhang2023huatuogpt}. They fail to offer critical diagnostic information or fully understand the nuances of a patient’s condition, which are essential for effective medical consultations.

To enhance the capabilities of Chinese medical LLMs, \citet{Yang_Zhao_Zhu_Zhou_Xu_Jia_Zan_2024} introduced Zhongjing, the first Chinese medical LLaMA-based LLM that implements an entire training pipeline. \citet{chen2023huatuogptiionestagetrainingmedical} open-sourced HuatuoGPT-II, a model that employs a one-stage training method to adapt to medical knowledge. ~\citet{liao2024ming} released MING-MOE, a Chinese medical LLM based on the Mixture-of-Expert architecture. 
However, the current works primarily focus on improving the performance of LLMs in Chinese medical tasks by optimizing training methodologies, while \emph{overlooking an in-depth investigation into the processing of data}, thereby neglecting the potential ability to achieve similar objectives through data processing~\cite{zheng2025large}.

To advance the development of Chinese medical LLMs, we propose DPF-CM, a Data Processing Framework for Chinese Medical LLM Training and Deployment. 
DPF-CM 
consists of two main modules. 
The first module is the complete data processing pipeline during the Continued Pre-Training, Supervised Fine-Tuning, and Reinforcement Learning. In these stages, in addition to basic data processing operations,
we identify a common limitation in open-source medical datasets—the lack of well-structured instructions, which diminishes the model’s generalization
capabilities and instruction comprehension capabilities. Inspired by the chain-of-thought (CoT) approach, we introduce a context-learning strategy based on chained examples to generate high-quality, question-oriented instructions. Specifically, we link different examples together to form a step-by-step thinking and progressive refinement process, ensuring that later examples are of higher quality than earlier ones. 
Furthermore, we introduce an averaging algorithm based on multiple reward models to cleanse the noisy samples within the preference data. Specifically, we train different reward models using the preference data. Subsequently, the trained reward models are employed to score the preference data, and the average score is calculated as the final score for each data sample. We then exclude data samples with scores below or above predefined thresholds. 

The second module focuses on training data privacy protection during deployment. Current open-source medical model processing pipelines often overlook the issue of data privacy. However, in the medical field, data privacy breaches can severely compromise patients' right to privacy. 
To this end, we propose PPVD to prevent training data leakage during deployment.
Specifically, PPVD first divides the training data sample into two parts, using the first half as prompts to query our trained medical LLM. The model's outputs are then matched against the second half to identify high-risk samples. Subsequently, we store the embeddings of these high-risk samples in a high-risk vector database and construct a corresponding secure vector database. During deployment, if a user's prompt matches content within the high-risk vector database, we respond using the content from the corresponding secure vector database.

We conduct extensive evaluation experiments on datasets encompassing single-turn medical dialogues, multi-turn medical dialogues, medical benchmarks, and medical terminology explanations. Our experiments demonstrate that DPF-CM effectively enhances model accuracy. The model we trained achieves SOTA performance among open-source Chinese medical LLMs. Moreover, the extent of training data privacy leakage is reduced by 27\%.

\section{Related Work}
The rapid development of large Chinese medical models owes much to the release of large Chinese language models. Initially, researchers trained these models by performing instruction fine-tuning on large language models using medical data. For instance, DoctorGLM~\citep{xiong2023doctorglmfinetuningchinesedoctor} collected diverse Chinese and English medical dialogue datasets and fine-tuned the ChatGLM-6B~\citep{team2024chatglm} model using P-tuning~\citep{liu-etal-2022-p}, enabling it to handle Chinese medical consultations. Similarly, DISC-MedLLM~\citep{bao2023discmedllmbridginggenerallarge} used the Baichuan-base-13B~\citep{yang2023baichuan} model, performing instruction fine-tuning on over 470,000 medical data points. 

Researchers have discovered that relying solely on instruction fine-tuning is insufficient to make a qualified medical consultation assistant. Consequently, some models adopted a more comprehensive training process, encompassing continued pre-training, instruction fine-tuning, and reinforcement learning.
For example, 
Zhongjing~\citep{Yang_Zhao_Zhu_Zhou_Xu_Jia_Zan_2024} underwent pre-training on various medical datasets, followed by instruction fine-tuning on single and multi-turn dialogues and medical NLP tasks, and further used reinforcement learning to ensure professionalism and safety. HuatuoGPT-II~\citep{chen2023huatuogptiionestagetrainingmedical} proposed a unified domain adaptation protocol, merging the previous two-stage process of continued pre-training and instruction fine-tuning into a single-step procedure.

\section{Methods}

\begin{figure*}[t]
\centering
  \includegraphics[width=0.8\linewidth]{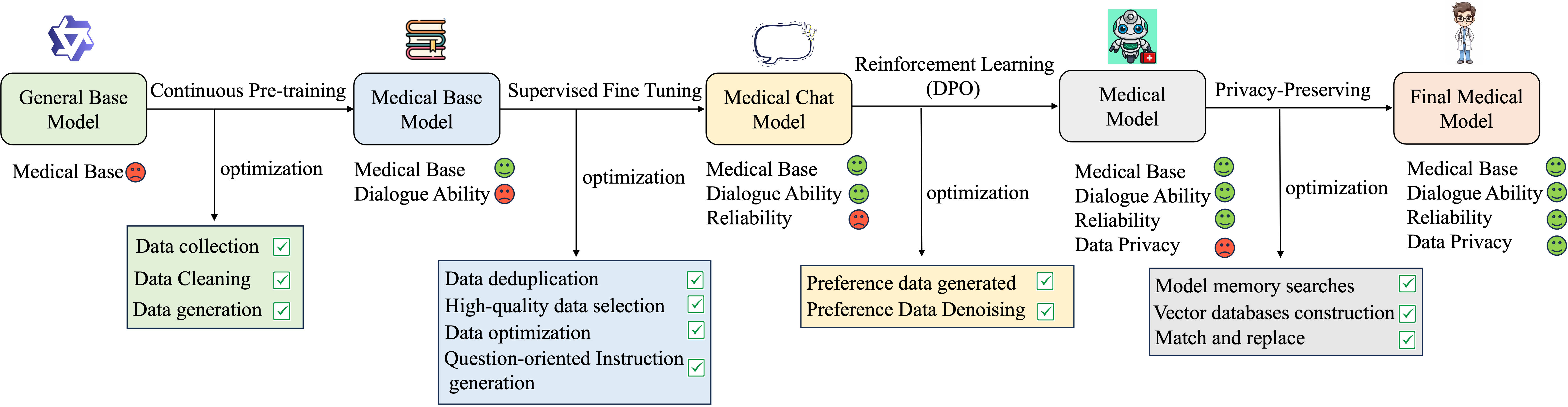}
  \caption {The overall flowchart of constructing DPF-CM. It encompasses the data processing methodologies throughout the
entire data lifecycle during the training and deployment of Chinese medical LLMs.
  }
  \vspace{-0.3cm}
  \label{fig1}
\end{figure*}

This section discusses the data processing pipeline of PF-CMLT in three training stages of Chinese medical LLM, including Continued Pre-training, SFT, and reinforcement learning. The comprehensive method flowchart is shown in Figure~\ref{fig1}.

\subsection{Processing of Continued Pre-Training Data}

For continued pre-training data, we first collect a vast amount of real medical pre-training corpus, which includes various types of medical data from different sources. We perform a comprehensive cleaning of the collected data. The cleaning strategies are as follows:

\ding{172} Too high character repetition rate or word repetition rate is considered to have content repetitive and needs to be filtered. \ding{173} Too high proportion of special characters indicates the presence of uncancellable page code or crawling artifacts, necessitating filtration.  \ding{174} Too high perplexity value suggests that the sentences are not fluent and require filtering.  \ding{175} An Insufficient number of words needs to be filtered. \ding{176} Noise characters in the text, such as HTML tags, are cleaned from the dataset. 

However, publicly available medical textbooks and popular science articles are scarce or require extensive web scraping. To address the data shortage, we employ the collected data as examples to guide the LLM in producing more data. The prompts used for data generation can be found in Appendix ~\ref{prompt_dg}. The statistics of the continued pre-training data can be found in Appendix ~\ref{pre-training-data}.

\subsection{Processing of SFT Data}

We first utilize the Minhash-LSH method~\cite{bai2023qwen} to deduplicate the data. 
Since most of the existing open-sourced Chinese medical dialogue data are derived from authentic patient-physician dialogues found on websites, this inevitably results in a substantial volume of low-quality data. Therefore, we use an LLM to select high-quality data, evaluating data quality across three dimensions: professionalism, safety, and fluency. The prompts we used can be found in Appendix ~\ref{prompt_s}. After selecting high-quality data, we further optimize the problematic datasets. We use the LLM to modify the data that needs improvement. The specific prompts can be found in Appendix ~\ref{prompt_o}.

\textbf{Question-oriented Instruction Generation:} Most existing medical data are in question-and-answer format, lacking instructions, which can weaken the model's generalization ability or instruction-following capability. To address this shortcoming, we have designed a chained example strategy to generation question-related instruction . 

We define the seed data as $D$ , which consists of $N$ samples in the format \{$(D_{1}^{ins}, D_{1}^{que}),....,(D_{N}^{ins}, D_{N}^{que})$\}. Each sample comprises a question and its corresponding instruction. Additionally, we define $\widetilde{D}$ as the data requiring instruction generation, containing $M$ samples in the format \{$(\widetilde{D_{1}^{que}},....,\widetilde{D_{M}^{que}})$\}, where each sample includes only the question part. Initially, the $D$ are sourced from the Chinese-medical-dialogue dataset\footnote{\url{https://huggingface.co/datasets/ticoAg/Chinese-medical-dialogue}} 


Few-shot is a commonly used prompting method for content generation, which can be formatted as $E_1 + ... + E_N + instruct +\widetilde{D_{1}^{que}} $, where
$E_N$ is selected from the $D$. 
By feeding this prompt into LLMs, the model will produce instructions for $\widetilde{D_{M}^{que}}$. However, this prompt construction method has two primary issues. 1) The quality of examples varies, which may cause the model to learn from low-quality examples, thus affecting the generation quality. 2) The examples are too segmented, preventing the model from efficiently learning relationships between different examples.

Inspired by the COT method, we link different examples together to enable a step-by-step thinking and progressive optimization process, allowing the model to produce better outputs along this chain. Based on this concept, we introduced chained examples. The prompt can be represented as $E_1 + instruct + E_2 + instruct^{*} + E_3 +... + E_N + instruct^{*}+\widetilde{D_{1}^{que}}$. 
To facilitate progressive optimization in chained examples, we score $D$ using LLMs, then we rank the examples from low to high quality as $E_1 < E_2,...,< E_N$. The prompts based on chained examples are shown in Appendix ~\ref{prompt_chained_examples}. By integrating different examples into chained examples, with a progressive optimization process, we encourage the LLM to contemplate why subsequent examples are better than preceding ones, yielding superior outputs.
The statistics of the processed SFT data can be found in Appendix ~\ref{pre-training-data}.

\subsection{Processing of Preference Data}

We use a two-stage processing method to generate high-quality medical preference data, which involves generating the data and removing the noise.

\begin{figure*}[!t]
\centering
  \includegraphics[width=0.67\linewidth]{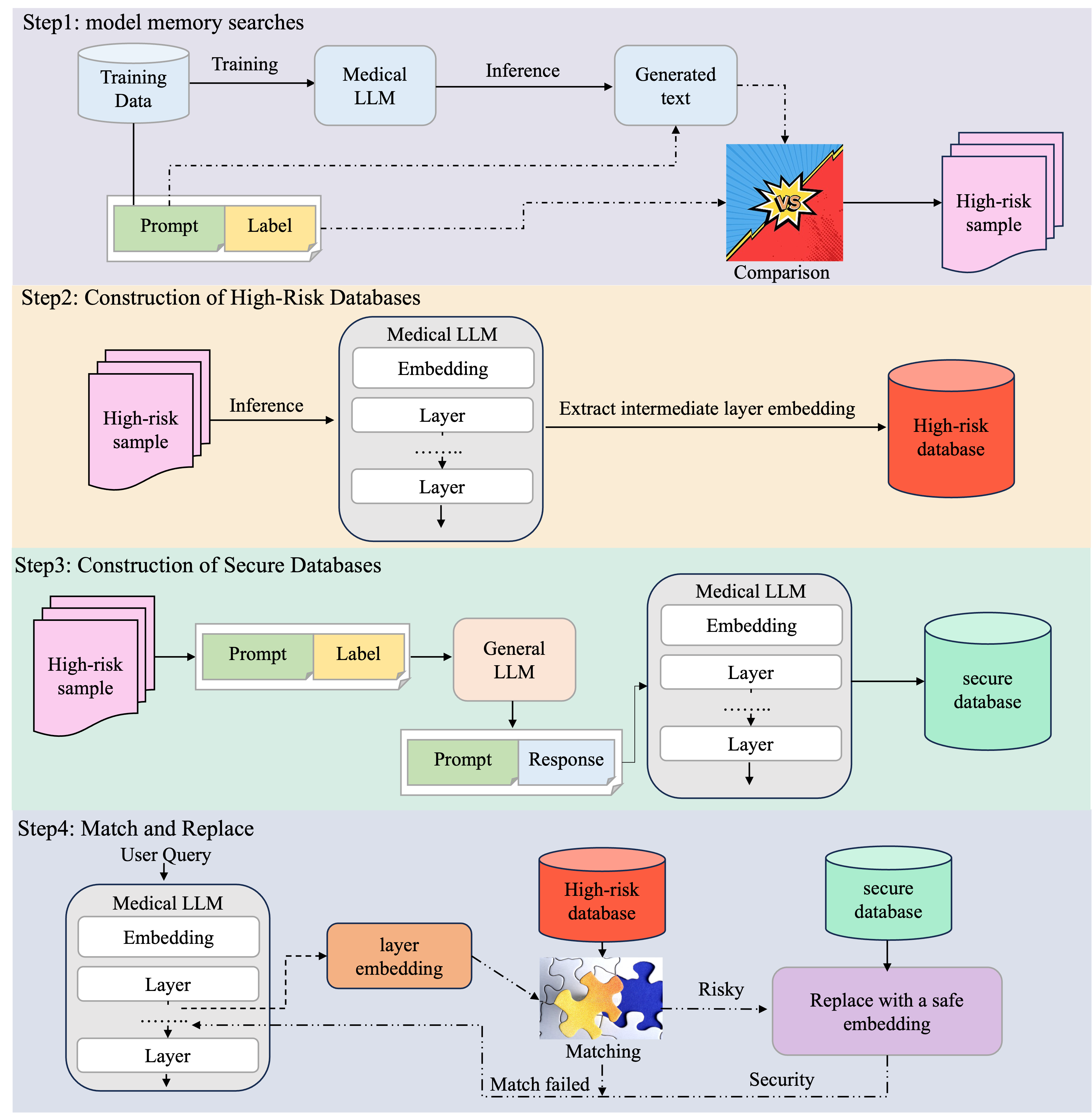}
  \caption {The PPVD algorithmic framework for data privacy protection in DPF-CM.}
  \label{fig3}
  \vspace{-0.3cm}
\end{figure*}

\textbf{Generate the Preference Data:} We randomly select 4,000 samples from the SFT dataset and an additional 2,000 samples from supplementary data as annotation data. The supplementary data comprises medical dialogue data not included in the SFT dataset, aiming to enhance the model's robustness. Subsequently, we generate five responses for each annotation using the SFT model. Finally, we employ three open-source LLMs to vote on the five responses to determine the optimal and least optimal replies. 

\textbf{Preference Data Denoising:} During the previous step of generating preference data, the following situations may arise:
1) The choice response and the reject response are contradictory. 2) The choice response and the reject response are essentially identical, lacking distinguishability. 3) The choice response and the reject response exhibit a certain degree of distinguishability. 4) In some cases, the distinguishability between the choice response and the reject response is excessively pronounced. Can we identify and remove data corresponding to the first and fourth scenarios? 

To this end, we use a denoising method for preference data, inspired by the loss function used in training reward models. During the training of reward models, treating the problem as a binary classification task yields the negative log-likelihood loss function: 
\begin{equation}
  \mathcal{L}(r_{\theta}) = -\mathbb{E}_{(x, y) \sim \mathcal{D}_{\text{rm}}} [\log \sigma(r_{\theta}(x, y_c) - r_{\theta}(x, y_r))]
\end{equation}
Here, $ {D}_{\text{rm}}$ denotes the preference dataset, consisting of \{$x^{i},y_c^i,y_r^i$\}. In the training of LLMs, $r_{\theta}$ is typically often initialized using the SFT model.

Because the term $r_{\theta}(x, y_c) - r_{\theta}(x, y_r)$ is used to measure the difference between the reward model's evaluations of different responses to the same query, we define this as the \textbf{preference distance}. We train five reward models using different random seeds. At the end of training, by aggregating the reward scores from these five reward models, we can compute the average preference distance for each pair of preference data. The computational formula can be found in Equation ~\ref{eq2}.
\begin{equation}\label{eq2}
  Dis^i\  =\  \frac{1}{N} \sum_{k=1}^{N}(r_{\theta_{k}}(x^i, y_c^i) - r_{\theta_{k}}(x^i, y_r^i))
\end{equation}
In the above formula, $i$ denotes the preference data sample, and $N$ represents the number of reward models.
If the preference distance trends of all reward models are either consistently less than zero or significantly large, we can identify the noisy data.

\section{Privacy Protection for Training Data}

Existing open-source medical model processing pipelines often neglect data privacy issues. However, in the healthcare field, data privacy breaches can severely infringe upon patients' right to privacy and significantly hinder the further development and deployment of Chinese medical LLMs. To facilitate breakthroughs in privacy protection for training data in Chinese medical LLMs, we propose a data privacy protection strategy based on vector databases. This scheme can be divided into four steps. In the following sections, we will provide a detailed description of these four steps. The algorithm flowchart can be seen in Figure ~\ref{fig3}.
The first aspect is identifying training data that is likely to be memorized by the large model. The specific steps are as follows:

\textbf{model memory searches:} The first aspect is identifying training data that is likely to be memorized by the LLM. We split a sample from the training set into two parts. The first part serves as a prompt for the attacker to input into the model, and the second part acts as the label to check whether the model's output matches this training sample. We use ROUGE-L to measure whether the model's output is sufficiently similar to the label. A higher similarity indicates that the model has a stronger "memory" of this training sample, thereby increasing the risk of privacy leakage. We consider samples with a similarity greater than a certain threshold to be memorized by the model. Subsequent protective measures mainly target these samples. 

\textbf{Construction of High-Risk Databases:} The second aspect is generating a high-risk embedding database by passing high-risk training data through the medical model and extracting intermediate embeddings. We use intermediate embeddings mainly to address potential leakage issues of the high-risk embedding database.

\textbf{Construction of Secure Databases:} We first divide high-risk samples into two components: Prompt and Label. The Prompt is then input into a general LLM, and the model's response is concatenated with the Prompt as a new sample. Finally, we pass the new sample through the medical model and extract intermediate embeddings to construct Secure Databases.

\textbf{Match and Replace:} The fourth step is the protection of training data. When the medical model is used for each user call, we compare the model's intermediate embeddings with the entries in the high-risk embedding database using cosine similarity. If the similarity exceeds a certain threshold, we return the intermediate results from the secure embedding database to the user. If the value is below the threshold, return the response directly.

\section{Experiment Setup}

\subsection{Datasets and Training Details}

Our model is based on Qwen2.5-7B~\citep{yang2024qwen2}. To assess the model's capability in medical dialogue, we incorporated specialized medical Q\&A data to simulate real doctor-patient interactions. This data includes both single-turn (huatuo26M~\citep{chen2023huatuogptiionestagetrainingmedical} and webMedQA~\citep{He2019}) and multi-turn medical conversations (CMtMedQA~\citep{Yang_Zhao_Zhu_Zhou_Xu_Jia_Zan_2024}). Concurrently, to evaluate the model's understanding and application of fundamental medical knowledge, we introduced a widely-used medical benchmark task (PLE, Ceval, CMB, CMMLU, and CMExam) and devised a medical terminology explanation task (medtiku\footnote{\url{https://www.medtiku.com/}}). 
The data we used is shown in Apppendix~\ref{data}. The training details can be found in Appendix ~\ref{training_details}.

\subsection{Baselines}
Our baseline includes two components: 1) a medical LLM that has not undergone data processing training to demonstrate the necessity of data processing optimization in DPF-CM; 2) various open-source Chinese medical LLMs, to illustrate that our trained model achieves SOTA performance among open-source models of the same size. We select four recently released, highly recognized open-source models in the Chinese medical field and conduct a comprehensive comparison with our Chinese medical model using the test dataset (HuatuoGPT-II~\citep{chen2023huatuogptiionestagetrainingmedical}, Zhongjing~\citep{Yang_Zhao_Zhu_Zhou_Xu_Jia_Zan_2024}, WiNGPT2~\citep{WiNGPT2}, and
ChiMed-GPT~\citep{tian-etal-2024-chimed}).
At the same time, we also compare our model with the most representative general LLMs, such as Qwen2.5-7B~\citep{yang2024qwen2} and GPT-4~\citep{achiam2023gpt}.

\subsection{Evaluation Metrics}

After studying the evaluation methods of other medical models, we adopted four evaluation approaches, assessing the model's performance based on the distinct characteristics of the aforementioned evaluation datasets, and aiming to comprehensively demonstrate the efficacy of our model. \textbf{(1) AI Evaluation}: We use GPT-4 as a tool to judge win, tie and loss rates. The prompt used for evaluation with GPT-4 is in Appendix~\ref{gpt4_prompt}. \textbf{(2) Similarity Evaluation:} We calculate the similarity between the model's output and the ground truth label to assess the quality of the generated responses. The similarity evaluation method used is the average of three metrics: 1/2 * (BERTScore + ROUGE\_L). \textbf{(3) Accuracy:}
For the medical benchmark, we use accuracy as an evaluation metric. \textbf{(4) Human Evaluation:} We hired three graduate students in the medical field as human evaluators to assess the quality of the generated text, on a daily payment basis. In cases of disagreement, we applied the majority voting principle. The evaluation dimensions are aligned with those used in AI assessment. 

\section{Experimental Results}

To fully demonstrate the performance of DPF-CM, we design the experiments from two perspectives: 1) We compare DPF-CM with models trained without any data preprocessing to highlight the effectiveness of DPF-CM. 2) We compare the Chinese medical LLM trained using DPF-CM with existing open-source Chinese medical LLMs of the same size to showcase the superiority of DPF-CM.

\renewcommand\arraystretch{0.8}
\begin{table}[!t]
\centering
\scalebox{0.75}{\begin{tabular}{cc|c}
\toprule
\multicolumn{2}{c}{Similarity Evaluation} & Ours. vs. Original\\
\midrule
Multi-turn dialogue & CMtMedQA & \textbf{0.833/0.000/0.167} \\
\midrule
\multirow{3}{*}{Single-turn dialogue} & All & \textbf{0.870/0.008/0.122} \\
~ & huatuo26M & \textbf{0.851/0.012/0.137} \\
~ & webMedQA & \textbf{0.889/0.003/0.107} \\
\midrule
Medical terminology & medtiku & \textbf{0.816/0.049/0.135} \\
\bottomrule
\end{tabular}}
\caption{Similarity evaluation between DPF-CM with models trained without any data preprocessing (expressed as '' Original'').  ``All'' refers to the evaluation results after merging huatuo26M and webMedQA. The above mathematical notation represents the win, tie, and loss rates format.}
\label{tab7}
\end{table}

\begin{table}[!t]
\centering
\scalebox{0.75}{\begin{tabular}{cc|c}
\toprule
\multicolumn{2}{c}{AI Evaluation} & Ours. vs. Original\\
\midrule
Multi-turn dialogue & CMtMedQA & \textbf{0.895/0.011/0.094} \\
\midrule
\multirow{3}{*}{Single-turn dialogue} & All & \textbf{0.913/0.003/0.084} \\
~ & huatuo26M & \textbf{0.904/0.005/0.091} \\
~ & webMedQA & \textbf{0.922/0.000/0.078} \\
\bottomrule
\end{tabular}}
\caption{AI evaluation between DPF-CM with models trained without any data preprocessing.}
\label{tab8}
\vspace{-0.2cm}
\end{table}

\begin{table}[!t]
\centering
\scalebox{0.75}{\begin{tabular}{cc|c}
\toprule
\multicolumn{2}{c}{Human} & Ours. vs. Original\\
\midrule
Multi-turn dialogue & CMtMedQA & \textbf{0.868/0.103/0.029} \\
\midrule
\multirow{3}{*}{Single-turn dialogue} & All & \textbf{0.851/0.121/0.028} \\
~ & huatuo26M & \textbf{0.847/0.125/0.028} \\
~ & webMedQA & \textbf{0.856/0.116/0.028} \\
\bottomrule
\end{tabular}}
\caption{Human evaluation between DPF-CM with models trained without any data preprocessing.}
\label{tab9}
\vspace{-0.5cm}
\end{table}

\begin{table}[!h]
\centering
\scalebox{0.75}{\begin{tabular}{c|cc}
\toprule
Multiple Choices & Ours. & Original \\
\midrule
PLE & \textbf{0.69} &0.62\\
Ceval & \textbf{0.80} &\textbf{0.80}\\
CMB  & \textbf{0.77} &0.74\\
CMMLU & \textbf{0.79} &0.78\\
CMExam & \textbf{0.73} &0.71\\
\bottomrule
\end{tabular}}
\caption{Multiple Choices Evaluation on DPF-CM with models trained without any data preprocessing.}
\label{tab10}
\vspace{-0.3cm}
\end{table}

\begin{table*}[!t]
\centering
\scalebox{0.72}{\begin{tabular}{cc|cccc}
\toprule
\multicolumn{2}{c}{Similarity Evaluation} & Ours. vs. HuatuoGPT-II & Ours. vs. Zhongjing & Ours. vs. ChiMed-GPT & Ours. vs. WiNGPT2 \\
\midrule
Multi-turn dialogue & CMtMedQA & \textbf{0.754/0.000/0.246} & \textbf{0.792/0.002/0.206} & \textbf{0.861/0.002/0.137} & 0.368/0.000/0.633 \\
\midrule
Multi-turn dialogue & CMtMedQA & \textbf{0.754/0.000/0.246} & \textbf{0.692/0.002/0.306} & \textbf{0.861/0.002/0.137} & \textbf{0.633/0.000/0.368} \\
\midrule
\multirow{3}{*}{Single-turn dialogue} & All & \textbf{0.638/0.008/0.354} & \textbf{0.619/0.014/0.367} & \textbf{0.580/0.013/0.407} & \textbf{0.625/0.010/0.365} \\
~ & huatuo26M & \textbf{0.664/0.008/0.328} & \textbf{0.586/0.016/0.398} & \textbf{0.586/0.008/0.406} & \textbf{0.612/0.008/0.380} \\
~ & webMedQA & \textbf{0.612/0.008/0.380} & \textbf{0.652/0.012/0.336} & \textbf{0.574/0.018/0.408} & \textbf{0.638/0.012/0.350} \\
\midrule
Medical terminology & medtiku & \textbf{0.760/0.003/0.237} & \textbf{0.738/0.002/0.260} & \textbf{0.787/0.005/0.208} & \textbf{0.754/0.001/0.245} \\
\bottomrule
\end{tabular}}
\caption{Similarity evaluation between our model and other Chinese medical LLMs.}
\label{tab11}
\vspace{-0.2cm}
\end{table*}

\begin{table*}[!h]
\centering
\scalebox{0.72}{\begin{tabular}{cc|cccc}
\toprule
\multicolumn{2}{c}{AI Evaluation} & Ours. vs. HuatuoGPT-II & Ours. vs. Zhongjing & Ours. vs. ChiMed-GPT & Ours. vs. WiNGPT2 \\
\midrule
Multi-turn dialogue & CMtMedQA & \textbf{0.391/0.487/0.122} & \textbf{0.701/0.256/0.043} & \textbf{0.988/0.010/0.002} & \textbf{0.621/0.313/0.066} \\
\midrule
\multirow{3}{*}{Single-turn dialogue} & All & \textbf{0.442/0.367/0.186} & \textbf{0.702/0.248/0.045} & \textbf{0.975/0.013/0.005} & \textbf{0.861/0.114/0.019} \\
~ & huatuo26M & \textbf{0.482/0.340/0.178} & \textbf{0.712/0.244/0.044} & \textbf{0.972/0.016/0.008} & \textbf{0.876/0.106/0.014} \\
~ & webMedQA & \textbf{0.402/0.394/0.194} & \textbf{0.692/0.252/0.046} & \textbf{0.978/0.010/0.002} & \textbf{0.846/0.122/0.024} \\
\bottomrule
\end{tabular}}
\caption{AI evaluation between our model and other Chinese medical LLMs.}
\label{tab12}
\vspace{-0.2cm}
\end{table*}

\begin{table*}[!h]
\centering
\scalebox{0.72}{\begin{tabular}{cc|cccc}
\toprule
\multicolumn{2}{c}{Human} & Ours. vs. HuatuoGPT-II & Ours. vs. Zhongjing & Ours. vs. ChiMed-GPT & Ours. vs. WiNGPT2 \\
\midrule
Multi-turn dialogue & CMtMedQA & \textbf{0.682/0.153/0.165} & \textbf{0.730/0.215/0.055} & \textbf{0.753/0.124/0.123} & \textbf{0.704/0.241/0.055} \\
\midrule
\multirow{3}{*}{Single-turn dialogue} & All & \textbf{0.496/0.326/0.178} & \textbf{0.624/0.253/0.123} & \textbf{0.798/0.101/0.101} & \textbf{0.857/0.042/0.101} \\
~ & huatuo26M & \textbf{0.483/0.216/0.301} & \textbf{0.683/0.152/0.165} & \textbf{0.875/0.021/0.104} & \textbf{0.784/0.052/0.164} \\
~ & webMedQA & \textbf{0.583/0.142/0.275} & \textbf{0.574/0.204/0.222} & \textbf{0.746/0.104/0.150} & \textbf{0.685/0.099/0.216} \\
\bottomrule
\end{tabular}}
\caption{Human evaluation between our model and other Chinese medical LLMs.}
\label{tab13}
\vspace{-0.2cm}
\end{table*}

\begin{table*}[!h]
\centering
\scalebox{0.72}{\begin{tabular}{cc|cccc}
\toprule
\multicolumn{2}{c}{Models} & \multicolumn{2}{c}{Ours. vs. Qwen2.5-7B-Instruct}  & \multicolumn{2}{c}{Ours. vs. GPT-4} \\
\midrule
\multicolumn{2}{c}{Metrics} & Similarity evaluation  & AI evaluation  & Similarity evaluation  & AI evaluation \\
\midrule
Multi-turn dialogue & CMtMedQA & \textbf{0.689/0.000/0.311} & \textbf{0.382/0.516/0.102} & \textbf{0.654/0.000/0.346} & \textbf{0.364/0.518/0.117} \\
\midrule
\multirow{3}{*}{Single-turn dialogue} & All & \textbf{0.613/0.007/0.380} & \textbf{0.354/0.523/0.119} & \textbf{0.540/0.007/0.453} & 0.143/0.505/0.352 \\
~ & huatuo26M & \textbf{0.662/0.010/0.328} & \textbf{0.358/0.518/0.124} & \textbf{0.550/0.006/0.444} & 0.158/0.516/0.326 \\
~ & webMedQA & \textbf{0.564/0.004/0.432} & \textbf{0.350/0.528/0.114} & \textbf{0.530/0.008/0.462} & 0.128/0.494/0.378 \\
\midrule
Medical terminology & medtiku & \textbf{0.863/0.003/0.134} & - & \textbf{0.842/0.000/0.158} & - \\
\bottomrule
\end{tabular}}
\caption{Similarity evaluation and AI evaluation between our model and general LLMs.}
\label{tab14}
\vspace{-0.2cm}
\end{table*}

\begin{table*}[!h]
\centering
\scalebox{0.72}{\begin{tabular}{c|cccccccc}
\toprule
Multiple Choices & HuatuoGPT-II & Zhongjing & ChiMed-GPT & WiNGPT2 & GPT-4 & Qwen2.5-7B-instruct & Ours. \\
\midrule
PLE & 0.47 & 0.31 & 0.48 & 0.42 & 0.69 & 0.61 & \textbf{0.69} \\
Ceval & 0.62 & 0.53 & 0.68 & 0.57 & 0.73 & 0.75 & \textbf{0.80} \\
CMB & 0.60 & 0.52 & 0.61 & 0.47 & 0.68 & 0.72 & \textbf{0.77} \\
CMMLU & 0.59 & 0.51 & 0.52 & 0.49 & 0.73 & 0.75 & \textbf{0.79} \\
CMExam & 0.65 & 0.55 & 0.53 & 0.51 & 0.68 & 0.69 & \textbf{0.73} \\
\bottomrule
\end{tabular}}
\caption{Multiple Choices Evaluation on our model and other LLMs.}
\label{tab15}
\vspace{-0.2cm}
\end{table*}

\subsection{Results of the Comparison Between DPF-CM and Raw Data}

Tables ~\ref{tab7}, ~\ref{tab8}, ~\ref{tab9}, and ~\ref{tab10} present the comparison results between DPF-CM and models trained without any data preprocessing. From the four tables, we can conclude that DPF-CM achieves significant performance improvements across all evaluation metrics. Particularly in medical dialogue tasks and medical terminology explanation tasks, the probability of DPF-CM winning the baseline reaches 85\%. These results demonstrate that positive domain-specific data preprocessing can effectively enhance the LLM's ability to learn domain knowledge, leading to superior performance.


\subsection{Results of the Comparison Between DPF-CM and Open-source Model}

The results of the comparison between DPF-CM and the other open-source LLMs in the Chinese medical fields are shown in Table \ref{tab11} and Table~\ref{tab12}. DPF-CM performs better than other open-source medical models in both metrics and datasets. The performance of DPF-CM in the medical benchmark is shown in Table~\ref{tab15}. DPF-CM exceeds all comparison models. Besides, DPF-CM outperforms all other models on the medical terminology task, reflecting the professionalism of the output content of DPF-CM. The results of human evaluation are shown in Table~\ref{tab13}, which show a similar trend to those of other evaluations. Comprehensive experiments demonstrate that our model has achieved the best open-source Chinese medical LLM among models of the same size.

\section{Ablation Study}

\subsection{Ablation Study on Training Data Processing}

To demonstrate the precision improvement contributed by each data optimization strategy at different training stages, we conduct various ablation experiments for validation. \emph{Due to space limitations in the main paper, we present two additional experiments in the appendix:} \emph{1) We explore the performance of cleaning and generation of pre-training data. The experimental results are shown in Appendix ~\ref{as_1}.} \emph{2) We explore the performance of selection and optimization of SFT Data. The experimental results are shown in Appendix ~\ref{as_2}.}

\textbf{Question-oriented Instruction Generation for SFT Data:} We conduct an ablation study on the question-oriented instruction generation for SFT data. 
Table~\ref{tab18} presents a comparison between the chained-example-based instruction generation method and a model trained without any instruction data. From the comparative experiments, we can conclude that the use of question-oriented instruction generation helps improve the model's accuracy, with more significant performance gains observed in multi-turn dialogues compared to single-turn dialogues. Table~\ref{tab18_more} compares the chained-example-based instruction generation approach with a general few-shot instruction generation method. The results show that our method outperforms the general approach, demonstrating that it is more effective in guiding the model to generate high-quality, question-oriented instructions.

\textbf{Denoising of Preference Data:} Table ~\ref{tab19} presents the ablation experiments related to the denoising of preference data. From the table, we can conclude that denoising preference data leads to a certain level of performance improvement, whether for multi-turn dialogue tasks, single-turn dialogue tasks, medical terminology explanations, or the medical benchmark. This indicates that removing noise from preference data during the DPO training phase can yield significant benefits. In other words, the acquisition of a high-quality set of preference data may significantly influence the accuracy of the DPO model.

\begin{table}[!t]
\centering
\scalebox{0.59}{\begin{tabular}{cccc}
\toprule
\multicolumn{2}{c}{QA} & AI Evaluation & Similarity Evaluation \\
\midrule
Multi-turn dialogue & CMtMedQA & \textbf{0.298/0.491/0.199} & \textbf{0.594/0.004/0.402} \\
\midrule
\multirow{3}{*}{Single-turn dialogue} & All & \textbf{0.323/0.397/0.271} & \textbf{0.504/0.017/0.407} \\
~ & huatuo26M & \textbf{0.328/0.394/0.272} & \text{0.492/0.010/0.498} \\
~ & webMedQA & \textbf{0.318/0.400/0.270} & \textbf{0.516/0.024/0.460} \\
\midrule
Medical terminology & medtiku & - & \textbf{0.564/0.014/0.422} \\
\midrule
\multicolumn{2}{c}{Multiple Choices} & \multicolumn{2}{c}{\textbf{0.748/0.745} (Accuracy)} \\
\bottomrule
\end{tabular}}
\caption{Ablation study of question-oriented instruction generation for SFT. Comparison between the chained-example-based instruction generation method and a model trained without any instruction data.}
\label{tab18}
\vspace{-0.1cm}
\end{table}

\begin{table}[!t]
\centering
\scalebox{0.59}{\begin{tabular}{cccc}
\toprule
\multicolumn{2}{c}{QA} & AI Evaluation & Similarity Evaluation \\
\midrule
Multi-turn dialogue & CMtMedQA & \textbf{0.256/0.446/0.218} & \textbf{0.444/0.137/0.419} \\
\midrule
\multirow{3}{*}{Single-turn dialogue} & All & \textbf{0.294/0.431/0.274} & \textbf{0.468/0.136/0.397} \\
~ & huatuo26M & \textbf{0.327/0.406/0.267} & \textbf{0.512/0.035/0.453} \\
~ & webMedQA & 0.262/0.456/0.282 & \textbf{0.423/0.236/0.341} \\
\midrule
Medical terminology & medtiku & - & \textbf{0.441/0.186/0.373} \\
\midrule
\multicolumn{2}{c}{Multiple Choices} & \multicolumn{2}{c}{0.748/0.748 (Accuracy)} \\
\bottomrule
\end{tabular}}
\caption{Ablation study of question-oriented instruction generation for SFT. Compares our chained-example-
based instruction generation approach with a general few-shot instruction generation method. }
\label{tab18_more}
\vspace{-0.1cm}
\end{table}

\begin{table}[!t]
\centering
\scalebox{0.59}{\begin{tabular}{cccc}
\toprule
\multicolumn{2}{c}{QA} & AI Evaluation & Similarity Evaluation \\
\midrule
Multi-turn dialogue & CMtMedQA & \textbf{0.255/0.623/0.122} & \textbf{0.615/0.000/0.386} \\
\midrule
\multirow{3}{*}{Single-turn dialogue} & All & \textbf{0.448/0.278/0.252} & \textbf{0.577/0.018/0.405} \\
~ & huatuo26M & \textbf{0.464/0.268/0.252} & \textbf{0.574/0.018/0.408} \\
~ & webMedQA & \textbf{0.432/0.288/0.252} & \textbf{0.576/0.018/0.406} \\
\midrule
Medical terminology & medtiku & - & \textbf{0.597/0.006/0.397} \\
\midrule
\multicolumn{2}{c}{Multiple Choices} & \multicolumn{2}{c}{\textbf{0.753/0.752} (Accuracy)} \\
\bottomrule
\end{tabular}}
\caption{Ablation study of denoising of preference data.}
\label{tab19}
\vspace{-0.3cm}
\end{table}

\subsection{Experiments on Different Base Models}

The experiments presented earlier were based on Qwen2.5-7B. Readers may wonder whether the performance of DPF-CM still outperforms existing open-source Chinese medical LLMs when replacing the base model. To address this concern, we conducted supplementary experiments. Zhongjing is a well-known Chinese medical LLM that has been fine-tuned through a complete training process. Based on this, we replaced the base model in DPF-CM with the same model used by Zhongjing (Ziya-LLaMA-13B-v1) and carried out comparative experiments to verify the effectiveness of the DPF-CM. The detailed experimental results are presented in Table~\ref{tab20} below. As shown in the table, our strategies remain effective even after switching the base model.

\begin{table}[!h]
\centering
\scalebox{0.59}{\begin{tabular}{cc|cc}
\toprule
\multicolumn{2}{c}{Metrics} & Similarity Evaluation & AI Evaluation  \\
\midrule
Multi-turn dialogue & CMtMedQA & \textbf{0.671/0.002/0.327} & \textbf{0.657/0.230/0.113}  \\
\midrule
\multirow{2}{*}{Single-turn dialogue} & huatuo26M & \textbf{0.681/0.012/0.307} & \textbf{0.643/0.140/0.217} \\
~ & webMedQA & \textbf{0.636/0.011/0.353} & \textbf{0.666/0.187/0.147} \\
\midrule
Medical terminology & medtiku & \textbf{0.7079/0.0047/0.2874} & -  \\
\midrule
\multicolumn{2}{c}{Multiple Choices} & \multicolumn{2}{c}{\textbf{0.617/0.484} (Accuracy)} \\
\bottomrule
\end{tabular}}
\caption{Ablation study on different base models.}
\label{tab20}
\vspace{-0.3cm}
\end{table}

\begin{table}[!ht]
\centering
\scalebox{0.74}{\begin{tabular}{ccccc}
\toprule
AI Evaluation & Datset & Original vs. Privacy-Safe\\
\midrule
Multi-turn dialogue & CMtMedQA & 0.152/0.693/0.155 \\
\midrule
\multirow{2}{*}{Single-turn dialogue} & huatuo26M & 0.211/0.580/0.209 \\
~ & webMedQA & 0.255/0.490/0.251 \\
\bottomrule
\end{tabular}}
\caption{Ablation study of Privacy-Safe. Using AI evaluation.}
\label{tab21}
\vspace{-0.2cm}
\end{table}

\begin{table}[!ht]
\centering
\scalebox{0.74}{\begin{tabular}{ccccc}
\toprule
Similarity Evaluation & Datset & Original vs. Privacy-Safe \\
\midrule
Multi-turn dialogue & CMtMedQA & 0.503/0.002/0.495 \\
\midrule
\multirow{2}{*}{Single-turn dialogue} & huatuo26M & 0.494/0.000/0.506 \\
~ & webMedQA & 0.487/0.014/0.499 \\
\bottomrule
\end{tabular}}
\caption{Ablation study of Privacy-Safe. Using Similarity evaluation as the evaluation metric.}
\label{tab22}
\vspace{-0.5cm}
\end{table}

\subsection{Results of Data Privacy Protection}

We select 100,000 samples from SFT and preference data to test the feasibility of the privacy protection methods proposed in the DPF-CM framework. We use the questions from these 100,000 samples as prompts and the corresponding answers as labels, following the first step described in Section 4 to retrieve the memory texts of the LLM. We set the similarity threshold at 0.85. Through experimentation, we find that only 1,812 samples out of the 100,000 met the similarity criterion of greater than 0.85, accounting for just 1.81\% of the total samples. Next, we execute the remaining steps and then retest the previously identified high-risk samples of 1,812. \textbf{The experimental results showed that the average similarity of these samples decreased to 0.58, which is a reduction of approximately 0.27.} This indicates that our method effectively protects the privacy of the pre-training data. Even if an attacker obtains a portion of the training data sample elements, they are unable to reconstruct the complete training dataset. 
We compare the model utilizing our aforementioned security scheme with a model that does not employ security measures. The results are shown in Tables~\ref{tab21}, ~\ref{tab22}. The table shows that our method has almost no impact on the model's performance.

\section{Conclusion}

In this paper, we propose DPF-CM to explore the value of data to the Chinese medical model from the perspective of data processing. DPF-CM encompasses the optimization of the entire data lifecycle, including continuing pre-training data, SFT data, preference data, and the privacy protection of training data. Through numerous experiments, we demonstrate that the Chinese medical models trained with the data processed by DPF-CM can achieve performance comparable to SOTA open-source medical LLMs of the same size. Additionally, we validate the necessity of each step of data processing within DPF-CM.

\section*{Limitations}

Despite DPF-CM demonstrating good performance across multiple test datasets and showing strong potential in many tasks, DPF-CM still has several limitations that need to be addressed and improved in future work.

DPF-CM utilizes general large language models to generate pre-training data. However, these large models may produce inaccurate or ethically inappropriate content due to their limited understanding of medical knowledge. In the future, we need to further explore domain-specific methods for generating pre-training data in the medical field to improve the quality of the generated data.

To ensure data privacy and security, the PPVD algorithm requires storing the embeddings of private data in High-Risk Databases. If the amount of private data is substantial, this can lead to significant storage demands on these databases. Therefore, in the future, we should explore more lightweight feature representations as alternatives to embeddings.



\bibliography{custom}

\appendix

\section{The Prompts Used in the Paper}

\subsection{The Prompts Used for Pre-train Data Generation}
\label{prompt_dg}
\noindent\rule{7.8cm}{0.6pt}
    As an expert with a professional medical background, your task is to compile high-quality content for medical textbooks.
    
    The compilation must meet the following requirements:
    
    Professionalism:
    
    $\bullet$ Provide scientific and accurate medical knowledge.
    
    $\bullet$ Clearly and concisely explain complex medical concepts.
    
    Safety:
    
    $\bullet$ Avoid creating content that could cause harm or lead to ambiguity.
    
    $\bullet$ Adhere to medical ethical standards to ensure compliance.
    
    Fluency:
    
    $\bullet$ Ensure semantic coherence, with no logical errors or irrelevant information.
    
    $\bullet$ Use language that is easy to understand to enhance readability.
    
    Examples of Medical Textbook Content:
    
    Example 1: 
    
    $[$Example content$]$
    
    Example 2: 
    
    $[$Example content$]$ 
    
    Example 3: 
    
    $[$Example content$]$ 
    
    Output Format Requirements: 
    
    Your output must strictly follow the format below: 
    
    Compiled Medical Textbook Content: 
    
    (The compiled contents for the medical textbooks are displayed here.) 
\noindent\rule{7.8cm}{0.6pt}

\subsection{The Prompts Used for SFT Data Selection}
\label{prompt_s}

\noindent\rule{7.8cm}{0.6pt} 
    As an evaluator with a professional medical background, please score the following medical data, which consists of a question and an answer from patient dialogues. 
    
    Question: 
    
    $[$Question content$]$ 
    
    Answer:
    
    $[$Answer content$]$ 
    
    The scoring criteria should be prioritized in the following order: Professionalism, Safety, and Fluency. The specific definitions are as follows:
    
    Scoring Criteria: 
    
    Professionalism: 
    
    $\bullet$ Accurately understand patients' questions and provide relevant answers.
    
    $\bullet$ Clearly and concisely explain complex medical knowledge.
    
    $\bullet$ Proactively inquired about relevant patient information when necessary. 
    
    Safety:
    
    $\bullet$ Provide scientific and accurate medical knowledge.
    
    $\bullet$ Honestly acknowledge when lacking knowledge about certain topics.
    
    $\bullet$ Ensure patient safety by refusing to offer information or advice that may cause harm. 
    
    $\bullet$ Adhere to medical ethics and respect patients' choices. 
    
    Fluency: 
    
    $\bullet$ Ensure semantic coherence, with no logical errors or irrelevant information. 
    
    $\bullet$ Use language that is easy to understand to enhance readability. 
    
    $\bullet$ Sustain a friendly and enthusiastic attitude in responses.
    
    Note: 
    
    Scoring must be based on the importance hierarchy of Professionalism > Safety > Fluency. In cases of conflict, prioritize the former.
    
    Please provide a score from 1 to 10 based on the overall assessment.
    
    If the data has deficiencies, apply strict deductions to widen the score range as much as possible.
    
    Your output must strictly follow the format below:
    
    Score Result:
    
    This section should contain only the score.
    
    Reason:
    
    This section should contain only your reasoning.
\noindent\rule{7.8cm}{0.6pt}

\subsection{The Prompts Used for SFT Data Optimization}
\label{prompt_o}
\noindent\rule{7.8cm}{0.6pt}
    As an optimization assistant for medical text data, your task is to evaluate and optimize the following medical question-and-answer data. 
    
    Data:
    
    $[$Data content$]$ 
    
    The criteria should be prioritized in the following order: Professionalism, Safety, and Fluency. The specific definitions are as follows: 
    
    Criteria: 
    
    [The criteria are consistent with the previous prompt.]
    
    Note: 
    
    Firstly, you need to determine whether the given data exhibits issues as outlined in the Criteria. If no such issues are present, optimization is not required; otherwise, optimization is necessary. Should optimization be required, please proceed to optimize the provided data. The optimized data must not only meet the requirements specified in the Criteria but also satisfy the following two conditions: 
    
    $\bullet$ Ensure that the core intent of the original input remains unchanged.
    
    $\bullet$ Maintain the length within a reasonable range (±30\%). 
    
    Your output must strictly follow the format below: 
    
    Data Requires Optimization: 
    
    yes/no 
    
    Optimized Data: 
    
    If optimization is needed, output the optimized data here; otherwise, output null. 
    
    Reason: 
    
    This section should contain only your reasoning. 
\noindent\rule{7.8cm}{0.6pt}

\subsection{The Prompts Based on Chained Examples}
\label{prompt_chained_examples}
\noindent\rule{7.8cm}{0.6pt}
    As an expert with a professional medical background.

    Your task is to generate the corresponding instruction for the given question, based on the provided examples. Please note that the examples we provide are chain-of-thought examples, representing a progressive optimization process—that is, the later examples produce higher-quality instructions than the earlier ones. You should study this optimization process and apply it to generate a better instruction for the given question.

    *Chain-of-thought examples:*
    
    Example 1: 
    
    $[$Example content$]$

    Please refer to the aforementioned example to generate relevant instruction for the following question.
    
    Example 2: 
    
    $[$Example content$]$ 

    Please refer to the previous instruction generation process to generate a relevant instruction for the following question.
    
    Example 3: 
    
    $[$Example content$]$ 

    Please refer to the previous chain-of-thought examples to generate a relevant instruction for the following question.

    Question:
    
    [Question content]
\noindent\rule{7.8cm}{0.6pt}

\section{Statistics of Data during Training} \label{pre-training-data}

Tables ~\ref{tab3} and ~\ref{tab6} present the statistics of the pre-training data and SFT data after being processed by DPF-CM.

\begin{table}[h]
\centering
\scalebox{0.72}{\begin{tabular}{ccc}
\toprule
Dataset & Type & Size  \\
\midrule
Medical Books & Books & 1.5  \\
Generated Medical Books & Books & 58.8  \\
CMtMedQA & Medical Dialogues & 2.5  \\
ChatMed-Consult-Dataset & Medical Dialogues & 7.9  \\
DISC-Med-SFT & Medical Dialogues & 14.7  \\
MedDiag & Medical Dialogues & 35.1  \\
cMedQA-V2.0 & Medical Dialogues & 2.2  \\
huatuo-sft-train-data & Medical Dialogues & 6.7 \\
webMedQA & Medical Dialogues & 5.6 \\
Chinese-medical-dialogue-data & Medical Dialogues & 11.5 \\
ShenNong-TCM & Knowledge Graph & 2.5 \\
huatuo-knowledge-graph-qa &Knowledge Graph & 3.0 \\
CMExam & Examination Questions &1.5 \\
CMB-Exam & Examination Questions &1.8 \\
PromptCBLUE & NLP Task &2.7 \\
Popular science articles & Articles & 20.3 \\
Generated Popular science articles & Articles & 83.5 \\
Paper abstracts & Articles &10.4 \\
baike2018qa & General Corpus & 37.8 \\
webtext2019zh & General Corpus & 76.3 \\
wiki2019zh & General Corpus & 42.4 \\
\bottomrule
\end{tabular}}
\caption{Statistics and sources of continued pre-training data in DPF-CM. The unit for Size is ten million tokens. Crawler refers to the medical science articles obtained from relevant medical websites. The medical paper denotes the abstracts of the collected medical research papers.
 }
\label{tab3}
\end{table}


\section{The Prompt used for Evaluation with GPT-4} \label{gpt4_prompt}
\noindent\rule{7.8cm}{0.6pt}
As a medical professional evaluator, please evaluate the following two doctors' responses to the same medical question. 

Question:

[Question content]

Response 1:

[Response 1 content]

Response 2:

[Response 2 content]

The evaluation criteria are prioritized in the following order: Accuracy of the doctor's response, Safety, Fluency, and Conciseness. The specific definitions are as follows:

Evaluation Criteria:

1. Accuracy of the Doctor's Response: The doctor should accurately understand the patient's question and provide a scientific and accurate answer.

2. Safety: The doctor must adhere to laws, regulations, ethics, and professional standards when answering.

3. Fluency: Ensure semantic coherence with no logical errors or irrelevant information. Maintain a friendly and warm tone.

4. Conciseness: Clearly and concisely explain complex medical concepts. Avoid unnecessary redundancy in the dialogue.

Note:

The importance of the evaluation criteria is ordered as Accuracy > Safety > Fluency > Conciseness. In case of conflicts, the higher-priority criterion takes precedence.

Your output must strictly follow the format below:

Evaluation Result:

Based on the above criteria, judge the result of “Response 1” relative
to “Response 2”. Output as: Win, Lose, or Tie.

Reason: (only reasons for rating can be answered here).
\noindent\rule{7.8cm}{0.6pt}

\section{Supplementary Experimental Setup}

\subsection{Datasets for Evaluation} \label{data}

1) Single-turn dialogue. Huatuo-26M~\citep{chen2023huatuogptiionestagetrainingmedical} is currently a large Chinese medical question-and-answer dataset. This dataset contains over 26 million high-quality medical Q\&A pairs, covering various aspects such as diseases, symptoms, treatment methods, and drug information. 
webMedQA~\citep{He2019} is a real-world Chinese medical question-answering dataset collected from online health consultancy websites. We use the test data to evaluate the medical models.

\begin{table}[!t]
\centering
\scalebox{0.75}{\begin{tabular}{ccc}
\toprule
Dataset & Type &Size\\
\midrule
ChatMed-Consult-Dataset & Medical Dialogues  & 43.3k\\
DISC-Med-SFT & Medical Dialogues  &53.9k\\
MedDiag & Medical Dialogues  &54.4k\\
huatuo-sft-train-data & Medical Dialogues &66.6k\\
PromptCBLUE & NLP Task &20k\\
alpaca-zh &Daily Dialogues &20k\\
BelleGroup/multiturn\_chat\_0.8M &Daily Dialogues &20k\\
BelleGroup/train\_0.5M\_CN &Daily Dialogues &20k\\

\bottomrule
\end{tabular}}
\caption{Statistics and sources of SFT data after processing. Size refers to the number of samples.
 }
 \label{tab6}
\end{table}

2) Multi-turn dialogue. The CMtMedQA\footnote{\url{https://huggingface.co/datasets/zhengr/CMtMedQA}} test set includes 1000 items for evaluating the model's multi-turn dialogue ability.

3) Medical Benchmark. We extracted questions about the medical field from C-Eval~\citep{NEURIPS2023_c6ec1844}, CMMLU~\citep{li-etal-2024-cmmlu}, CMExam~\citep{NEURIPS2023_a48ad12d}, CMB~\citep{wang-etal-2024-cmb}, and part of the 2023 Chinese National Pharmacist Licensure Examination~\citep{chen2023huatuogptiionestagetrainingmedical}.
4) Medical terminology explanation. We crawled medical terms and specialized explanations on the internet ourselves. For example, from medtiku\footnote{\url{https://www.medtiku.com/}}.

\subsection{Training Details} \label{training_details}
Our model is based on Qwen2.5-7B~\citep{yang2024qwen2}, a versatile LLM with 7 billion parameters. The training process utilized 24 A800-80G GPUs in parallel. We employ full-parameter fine-tuning, and to balance training costs, we use bfp16 precision alongside ZeRO-3~\cite{rajbhandari2020zero} and gradient accumulation strategies. The length of a single response, including its history, is capped at 2048 tokens. We incorporate the AdamW~\citep{loshchilov2017decoupled} optimizer, a 0.1 dropout rate, and a cosine learning rate scheduler. The best-performing checkpoint is retained as the final model. We employ
LLaMA-Factory~\citep{zheng2024llamafactory} as the training platform and vLLM~\citep{kwon2023efficient} for inference. 
For the Minhash-LSH algorithm, we use a deduplication threshold of 0.8. For preference data selection, we remove the highest-scoring 10 percent and the lowest-scoring 10 percent of the data. We set the scoring threshold to 9 and the similarity threshold to 0.8. The models used in the paper for data generation, data scoring, data optimization, and instruction generation are all Qwen2.5-72B\citep{yang2024qwen2}.

\section{Supplementary Ablation Study}
\subsection{Cleaning and Generation of Pre-Training Data} 
\label{as_1}

Since the model after continued pre-training lacks dialogue capabilities, we evaluate it using a medical benchmark. The ablation experiments are shown in Table ~\ref{tab16}. From the table, we can conclude that cleaning and generating the pre-training data help improve the model's accuracy. This is mainly because it reduces the interference of noisy samples and expands the knowledge scope of medical data, enabling the model to learn more efficiently and broadly.

\begin{table}[!h]
\centering
\scalebox{0.8}{\begin{tabular}{c|cc}
\toprule
Multiple Choices & Pretrained & SFT \\
\midrule
PLE & \text{0.58} & \textbf{0.61}  \\
Ceval & \text{0.68} & \textbf{0.73}  \\
CMB & \text{0.70} & \textbf{0.71}  \\
CMMLU & \text{0.66} & \textbf{0.71}  \\
CMExam & \text{0.61} & \textbf{0.69}  \\
\bottomrule
\end{tabular}}
\caption{Ablation study of cleaning and generation of pre-training Data.}
\label{tab16}
\end{table}

\subsection{Selection and Optimization of SFT Data} 
\label{as_2}

We conduct an ablation study on the selection and optimization of SFT data. From the results in Table ~\ref{tab17}, we can conclude that high-quality data selection and optimization lead to significant improvements, particularly for dialogue tasks. This is primarily due to enhancements in the expressiveness of the dialogues and the correction of erroneous knowledge, among other factors.

\begin{table}[!ht]
\centering
\scalebox{0.60}{\begin{tabular}{cccc}
\toprule
\multicolumn{2}{c}{QA} & AI Evaluation & Similarity Evaluation \\
\midrule
Multi-turn dialogue & CMtMedQA & \textbf{0.428/0.458/0.114} & \textbf{0.629/0.000/0.371} \\
\midrule
\multirow{3}{*}{Single-turn dialogue} & All & \textbf{0.536/0.270/0.171} & \textbf{0.585/0.115/0.300} \\
~ & huatuo26M & \textbf{0.576/0.228/0.182} & \textbf{0.586/0.110/0.304} \\
~ & webMedQA & \textbf{0.496/0.312/0.160} & \textbf{0.560/0.120/0.320} \\
\midrule
Medical terminology & medtiku & - & \textbf{0.614/0.005/0.381} \\
\midrule
\multicolumn{2}{c}{Multiple Choices} & \multicolumn{2}{c}{\textbf{0.748/0.740} (Accuracy)} \\
\bottomrule
\end{tabular}}
\caption{Ablation study of selection and optimization of SFT data. The "Multiple Choices" row in the table represents the average performance across five tasks from the medical benchmark.}
\label{tab17}
\end{table}

\end{document}